\documentclass[conference]{IEEEtran}
\usepackage[latin9]{inputenc}
\usepackage{array}
\usepackage{float}
\usepackage{url}
\usepackage{multirow}
\usepackage{amsmath}
\usepackage{amssymb}
\usepackage{graphicx}
\usepackage{xcolor}
\usepackage{soul}
\usepackage{flushend}
\usepackage[flushleft]{threeparttable}
\usepackage[ruled,vlined]{algorithm2e}
\usepackage[unicode=true, bookmarks=false, breaklinks=false,pdfborder={0 0 1},backref=section,pdftex,pdfborder={0 0 0},colorlinks=true,allcolors=darkblue]{hyperref}
\definecolor{darkblue}{rgb}{0.0, 0.0, 0.55}

\usepackage{booktabs}

\makeatletter

\newtheorem{theorem}{\textbf{Theorem}}

\newtheorem{lemma}[theorem]{\textbf{Lemma}}

\newtheorem{definition}{\textbf{Definition}}

\definecolor{Mem}{rgb}{.75,1,0}
\sethlcolor{Mem}

\makeatother

\begin{document}

\title{Fast \& Efficient Learning of Bayesian Networks from Data: Knowledge Discovery and Causality}

\author{\IEEEauthorblockN{Sein Minn$^{1}$, Fu Shunkai$^{2}$}
\\
\IEEEauthorblockA{
$^{1}$ Independent Researcher, Canada \\
$^{2}$ Huaqiao University, China\\
}}
\maketitle
\begin{abstract}\small\baselineskip=9pt 

Structure learning is essential for Bayesian networks (BNs) as it uncovers causal relationships, and enables knowledge discovery, predictions, inferences, and decision-making under uncertainty. Two novel algorithms, FSBN and SSBN, based on the PC algorithm, employ local search strategy and conditional independence tests to learn the causal network structure from data. They incorporate d-separation to infer additional topology information, prioritize conditioning sets, and terminate the search immediately and efficiently. FSBN achieves up to 52\% computation cost reduction, while SSBN surpasses it with a remarkable 72\% reduction for a 200-node network. SSBN demonstrates further efficiency gains due to its intelligent strategy. Experimental studies show that both algorithms match the induction quality of the PC algorithm while significantly reducing computation costs. This enables them to offer interpretability and adaptability while reducing the computational burden, making them valuable for various applications in big data analytics. 
\end{abstract}

\section{Introduction}
Bayesian networks provide a compact and expressive framework for modeling causality by utilizing joint probability distributions (JPDs) and enabling causal inference. This allows for a systematic representation of causal relationships between variables and facilitates reasoning and decision-making under uncertainty. They consist of a directed acyclic graph (DAG) where nodes represent random variables and conditional probability distributions are assigned to variables based on their parents in the graph~\cite{pearl1988probabilistic,2014Structure}. However, learning the graph structure of BNs from data poses significant challenges to computation and data efficiency, even when the data is complete~\cite{de2011efficient,2015Accelerated,2016Accelerating}.

\begin{figure}[h]
\centering 
  \includegraphics[scale=0.25]{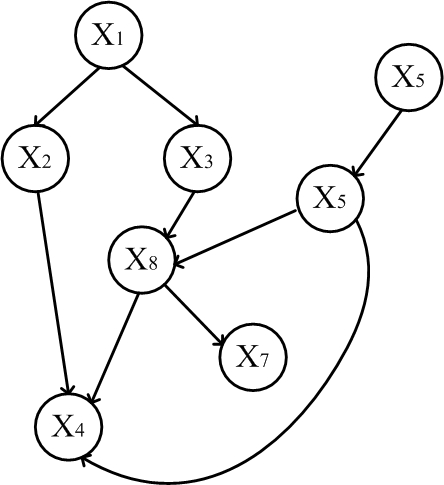}\\
  \caption{An example of Bayesian network structure.}\label{fig_bn}
\end{figure}

In the field of structure learning for Bayesian networks, existing algorithms can be broadly categorized into two approaches: constraint learning and search \& score methods. The constraint learning approach, outlined in the work by Spirtes, Glymour, and Scheines (SGS) \cite{spirtes2000causation}, involves performing a series of conditional independence tests on the available data. Based on the results of these tests, a BN is constructed that aligns with the observed dependencies among variables. In contrast, the search \& score approach, introduced by Cooper and Herskovits \cite{cooper1992bayesian}, focuses on finding a graph structure that maximizes a chosen scoring metric. This metric acts as a measure of the goodness of fit between the BN's structure and the provided data. The search \& score approach employs an iterative search process to explore and evaluate different graph structures based on the chosen scoring criterion~\cite{de2011efficient}. Both constraint learning and search \& score methods have their strengths and limitations. 

The constraint learning approach leverages statistical tests to directly identify the conditional independence relationships among variables, which allows for the construction of a BN that adheres to the observed dependencies in the data. On the other hand, search \& score methods provide flexibility in selecting a scoring metric that captures the desired properties of the BN. By searching through the space of possible graph structures, these methods aim to find the structure that optimally fits the given data according to the chosen scoring criterion~\cite{lam1994learning,heckerman1995learning}. The search \& score algorithms are known to be more computationally expensive compared to constraint-based learning algorithms, especially when dealing with large feature spaces~\cite{yu2020causality}. As a result, these algorithms can become practically infeasible for such scenarios. They use greedy strategies to explore the space of directed acyclic graphs (DAGs) and often rely on prior knowledge or assumptions to guide the search process. The selection of the appropriate approach depends on various factors, including the characteristics of the available data, the complexity of the BN structure, and the specific goals of the analysis. Researchers and practitioners often consider the trade-offs between time efficiency, accuracy in capturing dependencies, and the interpretability of the resulting BN structure when choosing among structure learning algorithms.

Both the constraint learning and search \& score approaches for structure learning in BNs are known to be computationally challenging. It has been proven that determining the optimal structure is an NP-hard problem \cite{chickering1994learning}, meaning that finding an exact solution becomes increasingly difficult as the number of variables grows. The best exact methods available exhibit exponential time complexity, which limits their applicability to small-scale problems with around 30 variables \cite{de2011efficient,fu2014towards,2016Efficient}. To handle larger networks, approximate procedures are often employed. However, these approximate methods can encounter difficulties in getting trapped in local maxima, leading to suboptimal structures. Despite this challenge, the quality of the learned structure is crucial for the accuracy of the resulting BN model. If the dependencies among variables are not correctly captured during the structure learning process, the estimated distribution may deviate significantly from the true distribution. Researchers and practitioners in the field continue to develop and refine algorithms that strike a balance between time efficiency and accuracy in structure learning. Various techniques, such as heuristics, optimization algorithms, and parallel computing, are being explored to improve the scalability and effectiveness of structure learning algorithms for BNs. 

This paper introduces two novel algorithms for learning the structure of Bayesian networks from data, both falling under the category of constraint-based learning methods. These algorithms utilize a series of local structure inductions, where conditional independence (CI) tests (by using $ {\chi}^2 $ test) are employed to determine the presence or absence of connections between pairs of nodes. To minimize the number of CI tests required, the algorithms utilize topology information inferred from previously conducted CI tests. This information is used to prioritize and sort future candidate CI tests. Tests that are more likely to reveal connections that should not exist are given higher priority, thereby avoiding unnecessary statistical tests. Additionally, considering that the CI tests are employed across various constraint-based learning algorithms for both inductions of Markov blankets and Bayesian networks, the proposed techniques can also be employed and may assist in accelerating the induction of the BN, as well as in the development of uni- and multi-dimensional BN classifiers. The effectiveness and efficiency of both algorithms are demonstrated through experiments conducted on synthetic and classical networks. The results highlight the algorithms' ability to accurately learn the structure of BNs while minimizing computational overhead. The remaining content is organized as follows: Section~\ref{sec:back}, notations, basic concepts of a Bayesian network, and its interpretability with causal reasoning are presented. We introduce these novel heuristics, while the specifications of the algorithms and the soundness of our approach with proofs are presented in Sections~\ref{sec:sl} and~\ref{sec:algo}, respectively. Experimental studies, prospective discussions, and the conclusion are presented in Sections \ref{sec:Exp}, and \ref{sec:con} respectively.

\section{Background and Related Works}\label{sec:back}
\subsection{Basic Knowledge}

Given a set of variables $\textbf{X}=(X_1,..., X_n)$, where $n \geq 1$, the Bayes theorem allows us to decompose the joint probability distribution of these variables into a product of conditional probabilities. Throughout the paper, a variable set by bold uppercase (e.g. $\textbf{X}$ ), a single variable by non-bold upper character (e.g. $X$), and their assignments by lowercase (e.g. $\textbf{X}=\textbf{x}, X=x$ ). A Bayesian network for $\mathbf{B}$ is defined as a pair $\mathbf{B = (G, \Theta)}$, where $G$ is a directed acyclic graph (DAG) with nodes corresponding to the random variables in $\mathbf{U}$. $\textbf{G=(V, A) }$ is a DAG,  $\textbf{G}$ in which $\textbf{V}$ corresponds to $\textbf{U}$, and an arc $\textbf{A}$ contains the connectivity information among $\textbf{V}$. If there exists a directed arc from node $X$ to node $Y$, we say that $X$ is a parent of $Y$, and $Y$ is a child of $X$. Nodes $X$ and $Y$ are considered neighbors if they are either parents or children of each other. The parameters $\theta$ represent the conditional probability distribution of each node $X \in \mathbf{B}$ given its parents. Parameter set $\mathbf{\Theta}$ contains $\mathbf{\Theta}_{x_i| Pa(x_i)} = P (x_i|\textbf{Pa}(x_i))$, each of which describes the conditional probability of the assignment $x_i$ given the assignment $\textbf{Pa}(x_i)$ to its parents. For those without parents, prior probability $P (x_i)$ is used instead. The joint probability distribution as represented by a BN, $\textbf{B}$ can be factorized below:

\begin{equation}\label{equ:cpd}
P(X_1,X_2,..,X_n) = \Pi_{i=1}^n P(X_i|\textbf{Pa}(X_i))
\end{equation}

Additionally, we define a path between nodes $X$ and $Y$ as any sequence of nodes such that consecutive nodes are connected by directed arcs, and no node appears in the sequence more than once. A directed path in a DAG is a path where nodes $(X_1, \ldots, X_n)$ satisfy $1 \leq i \leq n$ and $X_i$ is a parent of $X_{i+1}$. If there exists a directed path from node $X$ to node $Y$, we say that $X$ is an ancestor of $Y$ and $Y$ is a descendant of $X$. Nodes $X$, $Y$, and $Z$ form a v-structure if node $Z$ has two incoming arcs from $X$ and $Y$, but $X$ and $Y$ are not adjacent. This v-structure is denoted as $X \rightarrow Z \leftarrow Y$, where $Z$ is a collider. In the remainder of the paper, $|X|$ is used to denote the size of set $\mathbf{X}$. The notation $X \perp Y \mid \mathbf{Z}$ represents the conditional independence between $X$ and $Y$ given $\mathbf{Z}$, while $X \not\perp Y \mid \mathbf{Z}$ represents their conditional dependency.

\begin{definition}\label{df:mc}
\textbf{( Markov Assumption )} A node in a BN is independent of its non-descendant nodes, given its parents.
\end{definition}

\begin{definition}\label{df:FC}
\textbf{( Faithfulness Assumption )} A Bayesian Network $G$ and a joint distribution $P$ are faithful to one another iff. every (conditional) independence relation entailed by $P$ is also present in $G$.
\end{definition}

\begin{definition}\label{df:BP}
\textbf{( Blocked Path )} A path $S$ from $X$ to $Y$ is blocked by a set of nodes $\mathbf{Z}$ if one of the following conditions holds true: 
\begin{itemize}
    \item There is a non-collider node in $S$ belonging to $\mathbf{Z}$;
    \item There is a collider node Z on $S$ such that neither $Z$ non any of its descendants belong to $\mathbf{Z}$.
\end{itemize}
Otherwise, the path $S$ is known as \emph{unblocked} or \emph{active}.
\end{definition}

\begin{definition}\label{df:ds}
\textbf{( d-Separation )} Two nodes $X$ and $Y$ are d-separated by a set of nodes $\mathbf{Z}$ iff. every path from $X$ to $Y$ is blocked by $\mathbf{Z}$.
\end{definition}

Such a set $\mathbf{Z}$ is called a \emph{d-seperator} of $X$ from $Y$, denoted as $\mathbf{DS}(X,Y)$.

\begin{theorem} \label{the:1}
Two nodes are adjacent iff. there exists no set $\mathbf{Z}$ such that $X\perp Y\mid  \mathbf{Z}$ and $X, Y \not \in \mathbf{Z}$.
\end{theorem}

\begin{lemma} \label{lem:1}
Two nodes are not adjacent if there exists set $\mathbf{Z}$ such that $X\perp Y\mid  \mathbf{Z}$.
\end{lemma}

\subsection{Causality based Reasoning}

Methods of probabilistic reasoning have been developed to assist data scientists in understanding the dependencies among variables, allowing for the exploration of evidence needed for prognostic and diagnostic reasoning in decision-making systems. Bayesian networks have emerged as a valuable framework for representing the relationships between evidence and a target of interest, leveraging probabilistic properties to handle uncertainty in knowledge~\cite{taroni2004general,2015Accelerating,2016Algorithm}. Within a Bayesian network, probabilistic reasoning can be employed to account for direct causal influences originating from parent nodes that impact a specific variable $X_i$. This allows for the following lines of reasoning:
\begin{enumerate}
    \item In prognostic reasoning, the occurrence of an effect variable is influenced by a direct cause, represented by the parent variable $\textbf{Pa}(X_i)$. This means that if we have information about the cause variable, we can make predictions or inferences about the likelihood of the effect variable occurring. Prognostic reasoning is commonly used for prediction purposes, utilizing conditional probability tables associated with the target variable.
    \item Diagnostic reasoning in Bayesian networks involves inferring the probable causes or factors that contributed to the occurrence of an effect variable, based on the knowledge that the effect has already occurred. This type of reasoning involves reasoning against the causal direction and is useful for determining likely causes based on observed effects.
\end{enumerate}
\begin{figure}[h]
  \centering
  \includegraphics[scale=0.44]{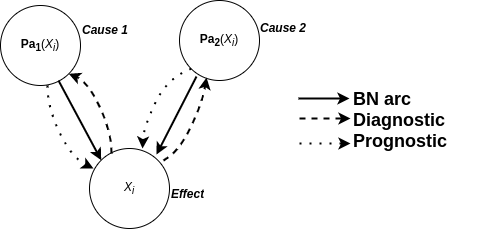}\\
  \caption{Illustration of prognostic and diagnostic reasoning in Bayesian networks.}\label{fig_cause_effect}
\end{figure}

\subsection{Related Works}
Constraint-based learning algorithms in Bayesian networks play a crucial role in uncovering the dependencies between variables. Algorithms of constraint-based learning rely on conditional independence (CI) tests (e.g. $ {\chi}^2 $ test), and they require enough samples to ensure the reliability of CI tests. At least five samples available per cell of the conditional probability table (CPT) is a widely accepted criterion~\cite{spirtes2000causation}. The majority of these algorithms assume faithfulness and utilize exact conditional independence (CI) tests, without requiring any specific ordering of nodes. Representative works in this category include the SGS algorithm and its updated version, the PC algorithm~\cite{spirtes2000causation}. The PC algorithm is specifically designed to avoid conducting high-number CI tests. Another algorithm, the IC algorithm~\cite{pearl1995theory}, shares similarities with PC algorithm and was proposed by Pearl et. al during the same period.
The GSBN algorithm, available in both original and random versions~\cite{margaritis1999bayesian}, follows the design of the SGS algorithm but relies on a sub-routine called GSMB to induce local neighborhoods. Although the time complexity of GSMB is polynomial, achieving acceptable results with this algorithm requires a large number of samples. There are a series of algorithms like GSMB those focus on only inducing the Markov blanket of the target variable. Those include Max-Min Parents and Children/MB~\cite{tsamardinos2003time}, HILTON-PC/MB~\cite{aliferis2003hiton}, Parents and Children based Markov Boundary~\cite{pena2007towards}, Iterative Parents and Children based Markov Blanket~\cite{fu2008fast}, and Markov Boundary search using the OR condition~\cite{rodrigues2008novel}. These algorithms mentioned aim to identify relevant features for the target variable using partially directed structures and CI tests, similar to the PC algorithm. However, their focus is solely on discovering the Markov blanket (MB) of a specific target variable in BNs for feature selection. It is worth considering that the proposed heuristics in this manuscript may be applicable to other structure learning scenarios and the induction of Markov blankets that also utilize CI tests. It is important to note that each category of constraint-based learning algorithms has its own strengths and limitations, and the choice of algorithm depends on factors such as available data, network complexity, and desired accuracy and time efficiency.

In order to improve the time efficiency of constraint-based learning methods for BNs, some approaches require the specification of node order, which entails stricter assumptions or expert judgment. The use of node order information can significantly reduce the search space and improve time efficiency. One early attempt in this direction was made by Wermuth and Lauritzen \cite{wermuth1982r}, but the Three-phase dependency algorithm (TPDA) by Cheng et. al \cite{cheng1997learning} is more widely recognized. TPDA relies on the assumption of ``monotone faithfulness" and achieves an acceptable time complexity of $O(n^4)$. Another algorithm, Polynomial Max-Min Skeleton (PMMS)~\cite{brown2005comparison}, achieves a similar time complexity with the same assumption. However, it has been shown by Chickering et al. \cite{chickering2003monotone} that the assumption of monotone faithfulness is not valid in many cases. Therefore, algorithms in this subgroup are not considered in this study.

Most of the algorithms mentioned earlier are focused on learning the skeleton or structure of the target networks, without determining the orientation of the edges. Additional rules or methods are required to establish the correct directionality of the edges. In general, the algorithms first identify the v-structures (or colliders) in the network, and then the remaining edges are arbitrarily oriented, as long as directed cycles are not introduced. The ultimate goal is to recover a model that accurately captures the underlying joint probability distribution (JPD). However, achieving a perfect match between the model and the true JPD is challenging due to two main factors: time efficiency and limited quality samples. 

Time efficiency refers to the computational challenges in accurately inferring the correct orientations of the edges, especially in larger networks. This complexity increases exponentially with the number of variables and can be a significant hurdle in achieving the ideal model. The limited number of available samples also poses a challenge. Insufficient data or bad quality data leads to incomplete information about the dependencies and relationships among variables, making it difficult to determine the orientations accurately. Algorithms alone cannot overcome this limitation since the underlying issue is the lack of comprehensive information. Considering these challenges, while it is desirable to recover a model that perfectly matches the underlying JPD, it is often not feasible in practice due to the aforementioned factors.

Nonetheless, the algorithms play a crucial role in approximating the true structure and providing valuable insights into the relationships among variables, even with the inherent limitations imposed by time efficiency and limited quality data samples.

\section{Structure Induction Algorithm Using Topology Information}\label{sec:sl}

Our work follows an overall strategy of recovering a BN from data samples through a series of local neighbor inductions. During the induction process, we employ a ``backward selection" approach, assuming that a node $X$ is connected to every other node $Y \in \mathbf{V} \setminus {X}$ and then eliminating false connections based on the presence of a d-separation ($\mathbf{DS}(X, Y)$). This backward selection technique, initially introduced in the PC algorithm, helps to keep the size of conditioning sets as small as possible. This, in turn, improves the reliability of conditional independence (CI) tests and enhances the overall learning quality. However, in previous works such as PC, all CI tests are treated equally and conducted randomly until a test reveals a conditional independence relation or no more tests remain. This approach can result in a significant number of fruitless CI tests. To address this issue, we propose two measures to quantify the likelihood of a test being able to uncover the desired CI relation. These measures are then utilized to develop two algorithms for the efficient recovery of BNs. 
\subsection{Search heuristics}\label{sec:sh}
In the context of the same dataset and two algorithms, A and B, if algorithm A achieves a more accurate outcome compared to B, it can be considered more ``data efficient". Among the existing constraint-based algorithms, the PC algorithm is recognized as the most data-efficient. The PC algorithm employs a strategy where it starts with an empty conditioning set and progressively expands it. This means that CI tests with smaller conditioning sets are given priority. However, while this strategy may be efficient to some extent, there is still room for improvement. One possible limitation is that it may not always be the most efficient approach. To address this, our work aims to enhance the PC algorithm by extracting and utilizing topology information that is hidden within the previously performed CI tests. By doing so, we can potentially improve the overall efficiency of the algorithm and further enhance its data efficiency.
\subsubsection{Search heuristic I}\emph{During the process of searching for neighbors of node $X$, we observe that certain nodes, such as node $Z$, may frequently appear in d-separators. Based on this observation, we can infer that node $Z$ has a higher likelihood of being selected as a candidate d-separator.}
\begin{figure}[htbp]
  \centering
  \includegraphics[scale=0.4]{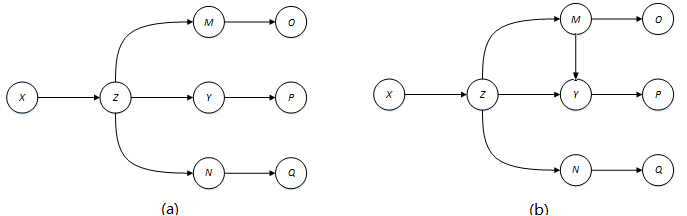}\\
  \caption{Illustration of selection diagrams depicting differences between heuristic I (a) and heuristic II (b).}\label{fig_d_separator}
\end{figure}

We can demonstrate the functionality of heuristic I using the network shown in Figure~\ref{fig_d_separator} (a). In order to determine whether there is no connection between nodes $X$ and $Y$, we need to identify the presence of a d-separator. Initially, the empty set is not considered a valid d-separator.

When examining d-separators of size 1, there are six potential sets to consider: ${Z}$, ${M}$, ${O}$, ${P}$, ${N}$, and ${Q}$. In the original PC algorithm, any of these sets may be chosen with equal probability. In the worst-case scenario, ${Z}$ could be the last set to be evaluated, resulting in a total of six CI tests. However, if we were able to select ${Z}$ as the first candidate, we could save the remaining five CI tests. This represents an 83\% reduction in computation compared to the worst-case scenario. Therefore, our objective is to identify the most likely candidate set to be a d-separator. To accomplish this, we introduce a quantitative measure called the \emph{relative d-separating ability (RDSA)}. For each node $X$, the RDSA of each node $Y \in \mathbf{V} \setminus {X}$ is initially set to 0. Whenever a d-separator $\mathbf{S}$ is discovered, the RDSA of each node $Z \in \mathbf{S}$ is updated as follows:
\begin{equation}
\forall Z \in \mathbf{S}, \mathbf{RDSA}^Z_X = \mathbf{RDSA}^Z_X + 1
\label{eq:DSA_X}
\end{equation}

\subsubsection{Search heuristic II}\emph{Sets with a higher number of nodes having a high RDSA value should be prioritized during the search process to improve the efficiency of identifying d-separators.}

Correspondingly we define the RDSA of a set $\mathbf{S}$ as summing up the RDSA of individuals as contained in $\mathbf{S}$ as below
\begin{equation}
\mathbf{RDSA}^{\mathbf{S}}_X = \sum_{Y\in \mathbf{S}} \mathbf{RDSA}^Y_X
\label{eq:DSA}
\end{equation}

We will demonstrate the effectiveness of these two measures by reusing the example mentioned above. After searching for the d-separator of size 1, ${Z}$ is identified as a d-separator, and $\mathbf{RDSA}^Z_X$ is updated to 1 according to Equation~\ref{eq:DSA_X}. During the search for the d-separator of $X$ and $M$, the set $\{\textbf{Z}\}$ is selected first due to its higher $\mathbf{RDSA}^{Z}_X$ value. Since $\{\textbf{Z}\}$ indeed serves as the d-separator for $X$ and $M$, the remaining CI tests can be disregarded, and $\mathbf{RDSA}^Z_X$ is incremented to 2. By continuing the search in this manner, the RDSA values of all $Y \in \mathbf{V}\setminus {X}$ are dynamically updated. Utilizing these values during the selection of candidate d-separators leads to a significant reduction in computational complexity, forming the foundation of the FSBN (Fast Search of BN) algorithm to be introduced in the next section.

\subsubsection{Search heuristic III}\emph{The relative contribution of a node $X$ in a d-separator $S$ is expected to be less significant compared to its occurrence in a smaller d-separator.}

Given the network shown in Figure~\ref{fig_d_separator}~(b), in order to d-separate $Y$ from $X$, we require a set that includes at least ${Z, M}$. However, the set ${Z}$ is sufficient to d-separate $X$ and $N$. To measure the RDSA in a more detailed manner, we propose the concept of \emph{weighted RDSA}, and its updating rule is as follows:
\begin{equation}
\forall Z \in \mathbf{S}, \mathbf{RDSA}^Z_X = \mathbf{RDSA}^Z_X + 1/|\mathbf{S}|
\label{eq:WRDSA_X}
\end{equation}

Equation \ref{eq:WRDSA_X} represents one possible implementation of heuristic III, which is utilized in the SSBN algorithm (Smart Search of BN). However, in SSBN, the calculation of RDSA for a set remains unchanged. It is hypothesized that there exist more advanced or smarter approaches for measuring the RDSA of a node or set, going beyond the simple strategy employed in FSBN. As a result, SSBN is proposed independently, even though it shares the same overall architecture as FSBN.

\section{Algorithm Specification}\label{sec:algo}

\subsubsection{\textbf{FSBN (Fast Search of BN)}} is constructed based on heuristic I and II, and it utilizes LSPC (Local Search of Parents/Child) to generate local neighbors for each node $X \in \mathbf{V}$. In this implementation, FSBN employs two global containers, $\mathbf{A}_{Ind}$ and $\mathbf{A}_{Del}$, to store induced disconnection and connection information in the format of $X-Y$.

The crucial component of FSBN is LSPC. When LSPC is called on node $T$, it assumes $T$ is connected to all nodes $X \in \mathbf{V}$, excluding pairs like $T-X$ that are stored in $\mathbf{A}_{Del}$. This prevents redundant analysis of pairs already known to be disconnected. Within LSPC, it starts with an empty set and iteratively grows it. Candidate d-separators of size $dss$ are generated using $\mathbf{GenSortDS}$, sorted in descending order based on their RDSA values. When a d-separator $\mathbf{S}$ is found (Line 11), the following steps are performed:
\begin{enumerate}
    \item $X$ is excluded from the neighbor set $\mathbf{V}_{CanPC}$ (Line 12);
    \item  $X-Y$ is removed from both $\mathbf{A}_{Can}$ and $\mathbf{A}_{Del}$ (Line 13 and 14); 
    \item The RDSA of each $Y \in \mathbf{S}$ is updated based on Equation \ref{eq:DSA_X} (Line 16);
    \item The analysis of node $X$ is terminated (Line 17), and the next node in $\mathbf{V}_{CanPC}$ is processed
\end{enumerate}

To save space, the explicit presentation of \textbf{GenSortDS} is omitted. It generates all possible subsets of $\mathbf{V}_{CanPC}$ of size $dss$, calculates their corresponding $\mathbf{RDSA}$ values according to Equation \ref{eq:DSA}, and sorts them in descending order.

$$\begin{array}{l}
\hline
  \textbf{FSBN}(\textbf{D}\colon Dataset)\hspace{10.09cm}\\
 01 \hspace{0.3cm} {\textbf{A}_{Ind}}= {\textbf{A}_{Del}}= \emptyset\\
 02 \hspace{0.3cm} for\hspace{0.1cm} (\forall X \in \textbf{V})\hspace{0.1cm} do\\
 03 \hspace{0.6cm} {\textbf{A}_{Can}} = \{X-Y | Y \in \textbf{V} \setminus \{X\}\} \setminus \textbf{A}_{Del}\\
 04 \hspace{0.6cm} {\textbf{A}_{Ind}}= {\textbf{A}_{Ind}} \cup\hspace{0.1cm} \emph{\textbf{LSPC}}(X,\textbf{D},{\textbf{A}_{Can}},{\textbf{A}_{Del}})\\
 05 \hspace{0.3cm} Apply\hspace{0.1cm} orientation\hspace{0.1cm} rules\hspace{0.1cm} on \hspace{0.1cm} {\textbf{A}_{Ind}}\\
 \hline
\end{array}$$

$$\begin{array}{l}

\hline
 \textbf{LSPC}(T,\textbf{D},{\textbf{A}_{Can}},{\textbf{A}_{Del}})\hspace{4.4cm}\\

 01 \hspace{0.3cm} dss=0; \\

 02 \hspace{0.3cm} {\textbf{V}_{CanPC(T)}}=\{ X| (T - X) \in {\textbf{A}_{Can}}\};\\

 03 \hspace{0.3cm} {\textbf{RDSA}^X_{T}}=0, \forall X \in \textbf{V}_{CanPC}; \\

 04 \hspace{0.3cm} while\hspace{0.1cm}(|{\textbf{V}_{CanPC}}|>dss)\hspace{0.1cm}\hspace{0.1cm}do \\

 05 \hspace{1.2cm} for\hspace{0.1cm} (\forall X \in {\textbf{V}_{CanPC(T)}})\hspace{0.1cm} do\\

 06 \hspace{1.8cm} if\hspace{0.1cm} (dss>0)\\

 07 \hspace{2.1cm} \textbf{SS} = \textbf{GenSortDS} ({\textbf{V}_{CanPC(T)}},dss,\textbf{RDSA}_{T});\\
 08 \hspace{1.8cm} else\\
 09 \hspace{2.1cm} \textbf{SS} = \emptyset;\\

 10\hspace{1.8cm} for\hspace{0.1cm}(\forall \hspace{0.1cm} \textbf{S} \in {\textbf{SS}})\hspace{0.1cm}do\\

 11 \hspace{2.3cm} if\hspace{0.1cm} (\emph{\textbf{I}}_D(T,X|\textbf{S}) \geq 1-\varepsilon)\hspace{0.1cm}then\\

 12 \hspace{2.8cm} {\textbf{V}_{CanPC(T)}}={\textbf{V}_{CanPC(T)}}\setminus \{X\};\\

 13 \hspace{2.8cm} {\textbf{A}_{Can}}= {\textbf{A}_{Can}} \setminus  \{T-X\};\\

 14 \hspace{2.8cm} {\textbf{A}_{Del}}= {\textbf{A}_{Del}} \cup  \{T-X\};\\

 15 \hspace{2.8cm} if\hspace{0.1cm} (|\textbf{S}|>0)\\

 16 \hspace{3.2cm} \forall \hspace{0.1cm} Y \in \textbf{S}, \hspace{0.1cm} \textbf{RDSA}^{Y}_{T}++; \\

 17 \hspace{2.8cm} break;\\

 18 \hspace{1.2cm} dss++;\\
 19 \hspace{0.3cm} return\hspace{0.1cm} {\textbf{A}_{Can}};\\
\hline
\end{array}$$

\subsubsection{\textbf{SSBN (Smart Search of BN)}} The overall architecture of SSBN is the same as FSBN, with the only difference being in Line 16 of LSPC, specifically the updating of RDSA for each node $X$ appearing in the d-separator $\mathbf{S}$. SSBN is designed based on heuristic III, thus it updates RDSA using Equation~\ref{eq:WRDSA_X}. The experimental study in the next section demonstrates that SSBN is significantly more efficient than FSBN.

\subsection{Proof of Soundness}\label{sec:pos}
The primary task of BN structure learning is to induce the skeleton of the target BN. Therefore, in our study, we focus on proving that both FSBN and SSBN will generate the correct skeleton. Both algorithms rely on the assumption of faithfulness and the use of accurate CI tests.
\begin{lemma} \label{lem:2}
LSPC will not remove $X-Y$ if it truly exists in the target network.
\end{lemma}

Assuming that a true edge $X-Y$ is included in $\mathbf{Z}_{Ind}$ implies that there exists a set of variables on which the conditional independence between $X$ and $Y$ holds. However, this contradicts Theorem~\ref{the:1}, which states that in a BN, there should be a direct edge between any two non-independent variables. Therefore, the inclusion of a true edge $X-Y$ in $\mathbf{Z}_{Ind}$ is not consistent with the underlying structure of a BN.

\begin{lemma} \label{lem:3}
The edge $X-Y$ determined to be false in \emph{LSPC($X$)} can be put into $\mathbf{A}_{Del}$, and be ignored in the call of \emph{LSPC($Y$)}.
\end{lemma}

According to the commutative law of probability, if variables $X$ and $Y$ are independent ($X \perp Y$), then it follows that $Y$ is also independent of $X$ ($Y \perp X$). This property holds in probabilistic independence relationships, allowing us to interchange the positions of variables without affecting their independence, and if $X\not \perp Y$ then $Y\not\perp X$.

\begin{lemma} \label{lem:4}
The edge $X-Y$ determined to be left in $\mathbf{A}_{Ind}$ by \emph{LSPC(X)} may actually be false.
\end{lemma}

Lemma~\ref{lem:1} states that nodes directly connected to $X$, including its parents denoted as $\mathbf{Pa}(X)$, will always be present in $\mathbf{V}_{CanPC}$. By the Markov condition, when $dss = |\mathbf{Pa}(X)|$, all non-descendant nodes will be removed from $\mathbf{V}_{CanPC}$ along with the corresponding edges. However, there may be cases where the edges between $X$ and its descendants fail to be removed from $\mathbf{A}_{Ind}$.

\begin{lemma} \label{lem:5}
False edge $X-Y$ left in $\mathbf{A}_{Ind}$ after the call of \emph{LSPC($X$)} will be removed during the call of \emph{LSPC($Y$)}.
\end{lemma}
Based on Lemma~\ref{lem:4}, we know that false edges only exist between node $X$ and it's descendant $Y$. It is important to note that $X$ can never be a descendant of $Y$ in the BN structure.

During the execution of \emph{LSPC(X)}, the set $\mathbf{V}_{CanPC}$ is constructed to include all nodes connected to $X$, excluding pairs like $X-Y$ that are stored in $\mathbf{A}_{Del}$. Since $X$ is not a descendant of $Y$, $Y$ will not be included in $\mathbf{V}_{CanPC}$.

Therefore, when \emph{LSPC(X)} is called, the false edge $X-Y$ will not be considered as a candidate d-separator, and it will not be added to $\mathbf{A}_{Ind}$. Consequently, the false edge $X-Y$ will be successfully removed from $\mathbf{A}_{Ind}$ during the execution of \emph{LSPC(X)}.

\begin{lemma} \label{lem:6}
True edges left in $\mathbf{A}_{Ind}$ after the call of \emph{LSPC(X)} will still be found in $\mathbf{A}_{Ind}$ after the call of \emph{LSPC(Y)}.
\end{lemma}

We will prove that FSBN and SSBN will produce the correct BN structure by considering the following points:

Lemma~\ref{lem:2} and Lemma~\ref{lem:6} guarantee that true edges will not be wrongly deleted during the execution of FSBN. Lemma~\ref{lem:2} ensures that true edges will be correctly identified as d-separators, while Lemma~\ref{lem:6} ensures that true edges will not be mistakenly removed from $\mathbf{A}_{Ind}$.

Lemma~\ref{lem:3} and Lemma~\ref{lem:5} demonstrate that false edges, if present in $\mathbf{A}_{Ind}$, will be successfully removed during the execution of FSBN. Lemma~\ref{lem:3} ensures that false edges between non-descendant nodes are removed during the LSPC process, while Lemma~\ref{lem:5} ensures that false edges between a node $X$ and its descendant $Y$ will be removed when \emph{LSPC(Y)} is called.
The search process in FSBN is applied to each node individually. This means that each node is considered as the starting point of the LSPC process, ensuring that the correct d-separators and edges are identified for each variable.

By considering these points, we can conclude that FSBN and SSBN will produce the correct BN structure. It ensures that true edges are preserved, false edges are removed, and the search is performed for each node individually. Therefore, these two algorithms are capable of recovering the correct BN based on the given data, under the assumptions of faithfulness and correct CI tests.

\section{Experimental Study}\label{sec:Exp}

\subsection{Settings}
Two classic well-known real networks, namely Asia and Alarm, are selected, along with two synthetic larger networks, 100\_130 and 200\_250, which were synthetically generated using the BN package of Weka. These networks are chosen for comparison purposes (Table \ref{tb_data}). To evaluate the performance and robustness of the proposed methods, we conducted experiments by randomly generating samples from each of the four networks. We varied the population sizes across a range of values, specifically from 1000 to 5000. For each population size, we generated 10-fold datasets, ensuring a diverse and representative set of samples in our experiment.

\begin{table}[h]
\caption{Description of data sets}\label{tb_data}
\centering
\begin{tabular}{lccc}
  \hline
  Data set(s)  & \# Instances & \# Nodes & \# Arcs\\
  \hline
  Asia & 1K to 5K & 8 & 8 \\
  Alarm & 1K to 5K  & 37& 46 \\
  100\_130 & 1K to 5K  & 100 & 130 \\
  200\_250 & 1K to 5K  & 200 & 250\\
  \hline\hline
\end{tabular}

\end{table}

Only PC algorithm is chosen for comparison considering: 
\begin{enumerate}
    \item Most well-known and theoretically sound \cite{kitson2023survey},  
    \item Most data efficient methods among constraint-based learning algorithms~\cite{yu2020causality},
    \item Monotone assumption required by a subgroup of algorithms like TPDA and PMMS is proved not practical~\cite{chickering2003monotone},
    \item FSBN and SSBN have the same assumptions as PC and all methods apply $ {\chi}^2 $ test as conditional independence test.
\end{enumerate}
All the algorithms are implemented within the Weka framework, and the PC algorithm implemented in Weka is used directly. For each experiment, we calculated and reported average performance metrics. These metrics provided an overall assessment of the network's performance across the different folds and population sizes. The evaluation is performed based on the following four dimensions:

\emph{Knowledge discovery-oriented learning quality}: Each edge is treated equally in knowledge discovery, regardless of the strength of the encoded dependency relation. The ideal measure for this task is the~\emph{Hamming distance}~\cite{hamming1950error}, which is defined as the number of actions required to transform the learned network into the target network. The allowed actions include adding, deleting, and reversing edges.

\emph{Density estimation-oriented learning quality}: This dimension aims to assess the quality of the global network, distinct from knowledge discovery, by considering the marginal likelihood of the training data over the parameter distribution for the candidate network. To measure this, the Bayesian Dirichlet likelihood-equivalent uniform joint distribution~\cite{buntine1991theory,liu2012empirical} \emph{BDeu score} is employed. The \emph{BDeu score} enables a comparison between the output network and the structure of the true underlying model, assuming that the real network is known. It provides a means to evaluate the fit of the candidate network to the training data in terms of the underlying model's structure.

\emph{Time efficiency}: The time efficiency is measured by calculating their computational complexity in this study and is derived from the summation of weights associated with each executed conditional independence test. Specifically, the weight assigned to the test $I_D(X, Y|\mathbf{Z})$ is defined as $2 + |\mathbf{Z}|$. This weight-based approach serves as a standard method for comparing conditional independence test-based algorithms~\cite{minn2014efficient}. It is justified by the observation that the runtime of a statistical test on a triplet $(X, Y |\mathbf{Z})$ is directly proportional to the size of the dataset and the number of variables involved. Notably, the time complexity of the test is $O(N (|\mathbf{Z}|+2))$ ( without exponential growth according to the number of variables, which a naive implementation might wrongly assume). The reason behind being able to construct all non-zero entries in the contingency table used by the test by examining each point in the dataset exactly once is that it can be done in a time proportional to the number of variables involved in the test. Specifically, the time required is proportional to the sum of the cardinality of the set {X, Y} and the set \textbf{Z}, which can be expressed as $|{X, Y} + \textbf{Z}| = 2 + |\textbf{Z}|$. This property stems from the nature of the conditional independence test and its relationship to the contingency table. This efficiency measure is machine-independent, allowing for the evaluation of computing costs across various observed quality levels. Synthetic data sets are utilized in this study to provide insights into the corresponding computational expenses under different quality conditions.

\emph{Influence of network density}: Increasing the density of the network results in a higher-dimensional solution space, which may transform a feasible problem into an infeasible one. Studying the influence of denser networks helps determine the effectiveness of the newly proposed search heuristics.

These four dimensions provide a comprehensive evaluation of the algorithms.

\subsection{Quality of Knowledge Discovery}
\begin{figure}[h]
  \centering
  \includegraphics[scale=0.3]{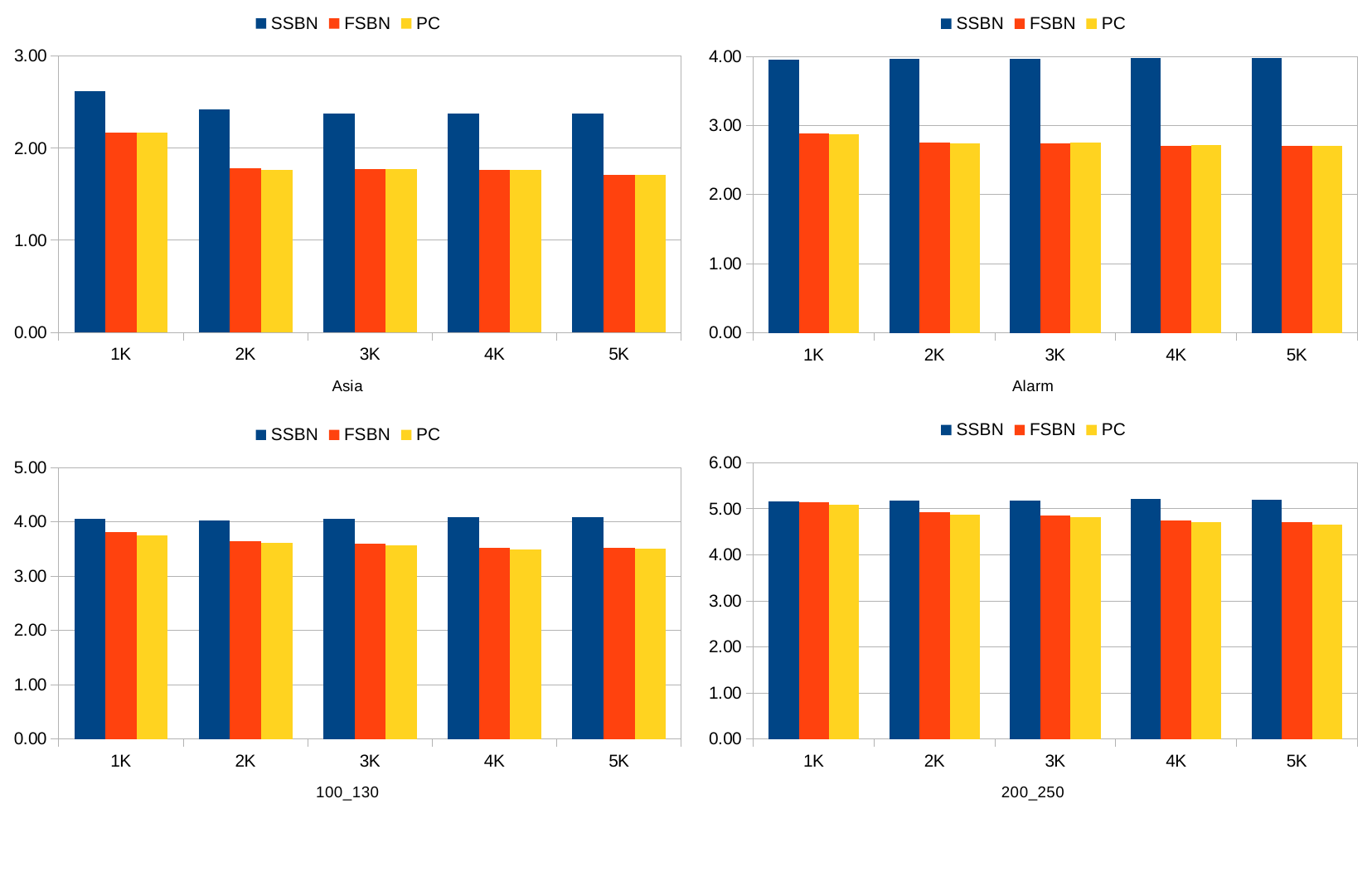}\\
  \caption{The Hamming Distance between the true networks and the induced networks by PC, FSBN, and SSBN (Note that the values on the Y-axis vary for each graph in this figure).}\label{fig_HD}
\end{figure}

Figure \ref{fig_HD} depicts the knowledge discovery quality of the PC, FSBN, and SSBN algorithms. It shows the number of actions required to transform the induced network into the true network. PC and FSBN exhibit similar performance, with low Hamming Distances indicating a close resemblance to the true network, and observed that FSBN achieves nearly the same quality of networks as the PC algorithm. SSBN shows slightly poorer performance compared to the other two algorithms, but it closes the gap with FSBN and PC as the network size increases. For instance, when applied to the Alarm network, the accuracy rate of SSBN is 91.37\% (3.97/46) (which means it requires correction of average 3.97 arcs in an alarm network with 46 arcs), but it improves to 96.86\% (4.09/130) and 97.92\% (5.20/250) for the 100\_130 and 200\_250 networks, respectively. Therefore, both FSBN and SSBN perform well enough in the knowledge discovery task.

\subsection{Quality of Density Estimation}

\begin{figure}[h]
  \centering
  \includegraphics[scale=0.3]{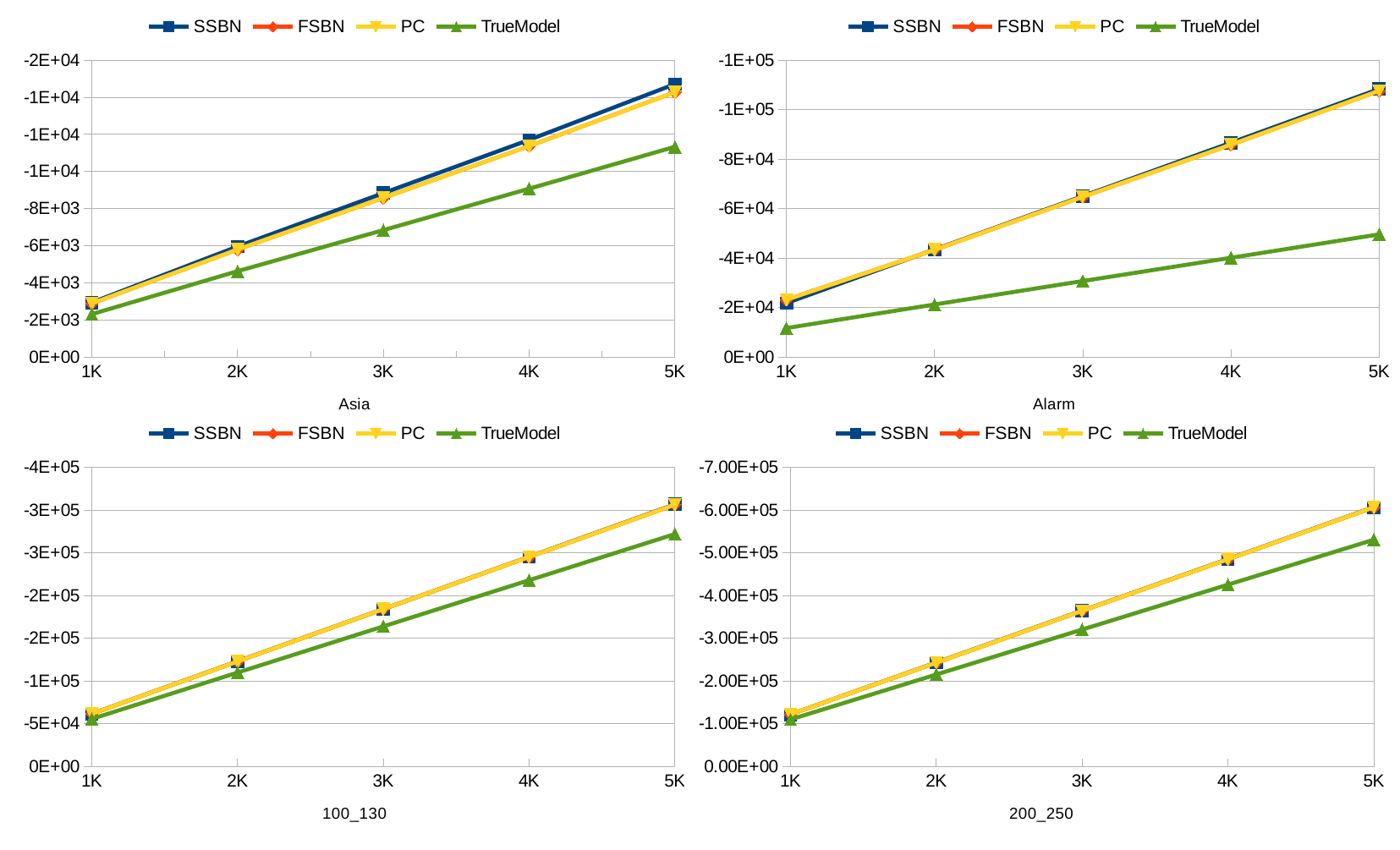}\\
  \caption{Quality of density estimation (\emph{BDeu score}) about PC, FSBN, SSBN, and true networks (Note that the values on the Y-axis vary for each graph in this figure).}\label{fig_Performance}
\end{figure}

Figure \ref{fig_Performance} illustrates the quality of the density estimation task achieved by PC, FSBN, and SSBN. The results show that all three algorithms perform comparably well across different sample sizes and network structures. This indicates that the newly proposed heuristics in FSBN and SSBN are effective in producing accurate density estimates. The similarity in performance among the algorithms further supports the conclusion drawn from the knowledge discovery task, highlighting the effectiveness of the proposed heuristics in producing satisfactory results.

\subsection{Time efficiency}
\begin{figure}[h]
  \centering
  \includegraphics[scale=0.3]{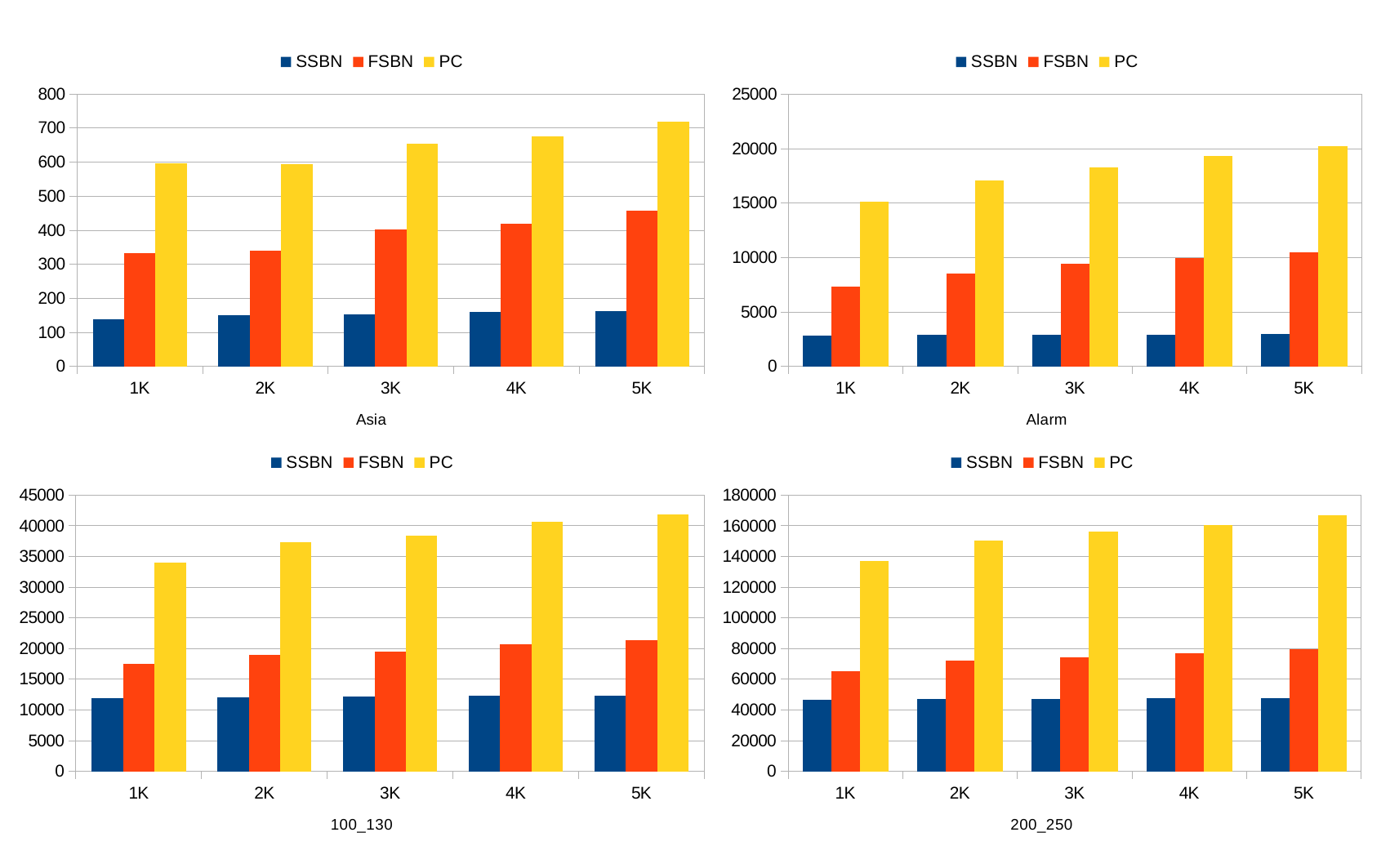}\\
  \caption{Computational complexity of PC, FSBN, and SSBN to induce given networks (Note that the values on the Y-axis vary for each graph in this figure).}\label{fig_time}
\end{figure}

Figure \ref{fig_time} showcases the relative time efficiency achieved by PC, FSBN, and SSBN. The following observations can be made:

\begin{enumerate}
    \item FSBN and SSBN demonstrate a clear advantage over the PC algorithm in terms of time efficiency. For instance, when applied to a dataset of 5,000 samples from the 200\_250 network, PC requires an average of 166,771 weighted CI tests. In contrast, FSBN and SSBN only require 79,446 and 47,569 tests, respectively, representing a reduction of 52.4\% and 71.5\% in the number of tests needed.
    \item SSBN exhibits greater time efficiency compared to FSBN, despite both algorithms sharing the same overall framework. For example, when applied to the 200\_250 network, SSBN requires 40.1\% fewer CI tests than FSBN. 
    \item With an increase in the number of samples, all three algorithms experience an increase in the number of CI tests required. However, the rate of increase is significantly slower for SSBN compared to the other two algorithms.
    
\end{enumerate}

These results demonstrate that both FSBN and SSBN offer improved time efficiency compared to the PC algorithm, with SSBN exhibiting the highest level of efficiency among the three algorithms, particularly as the number of samples increases.

\subsection{Influence of Network Density}
\begin{figure}[htbp]
  \centering
  \includegraphics[scale=0.28]{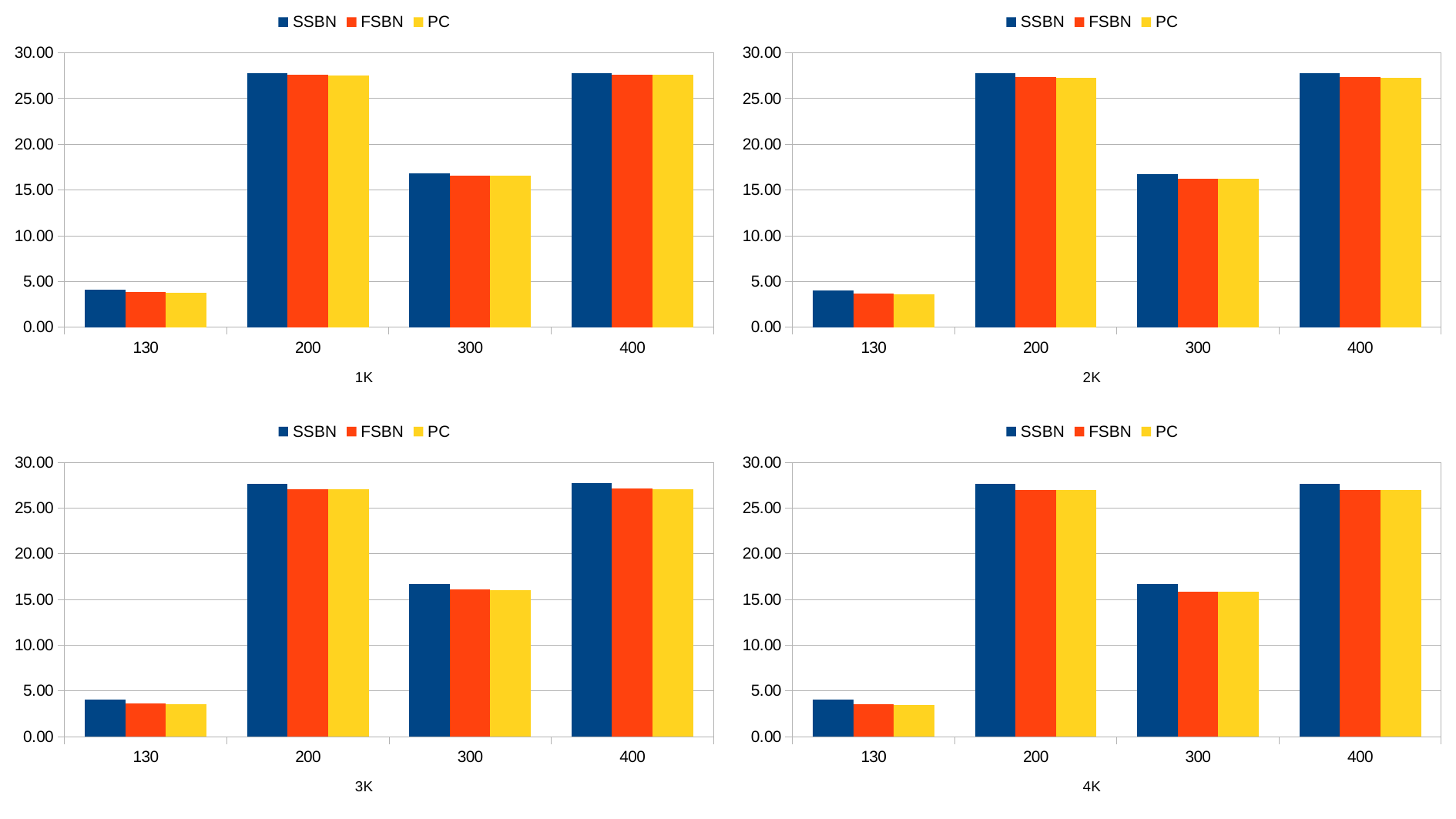}\\
  \caption{Hamming distance about PC, FSBN, and SSBN to true networks with various densities.}\label{figd_den_hamming}
\end{figure}

\begin{figure}[htbp]
  \centering
  \includegraphics[scale=0.31]{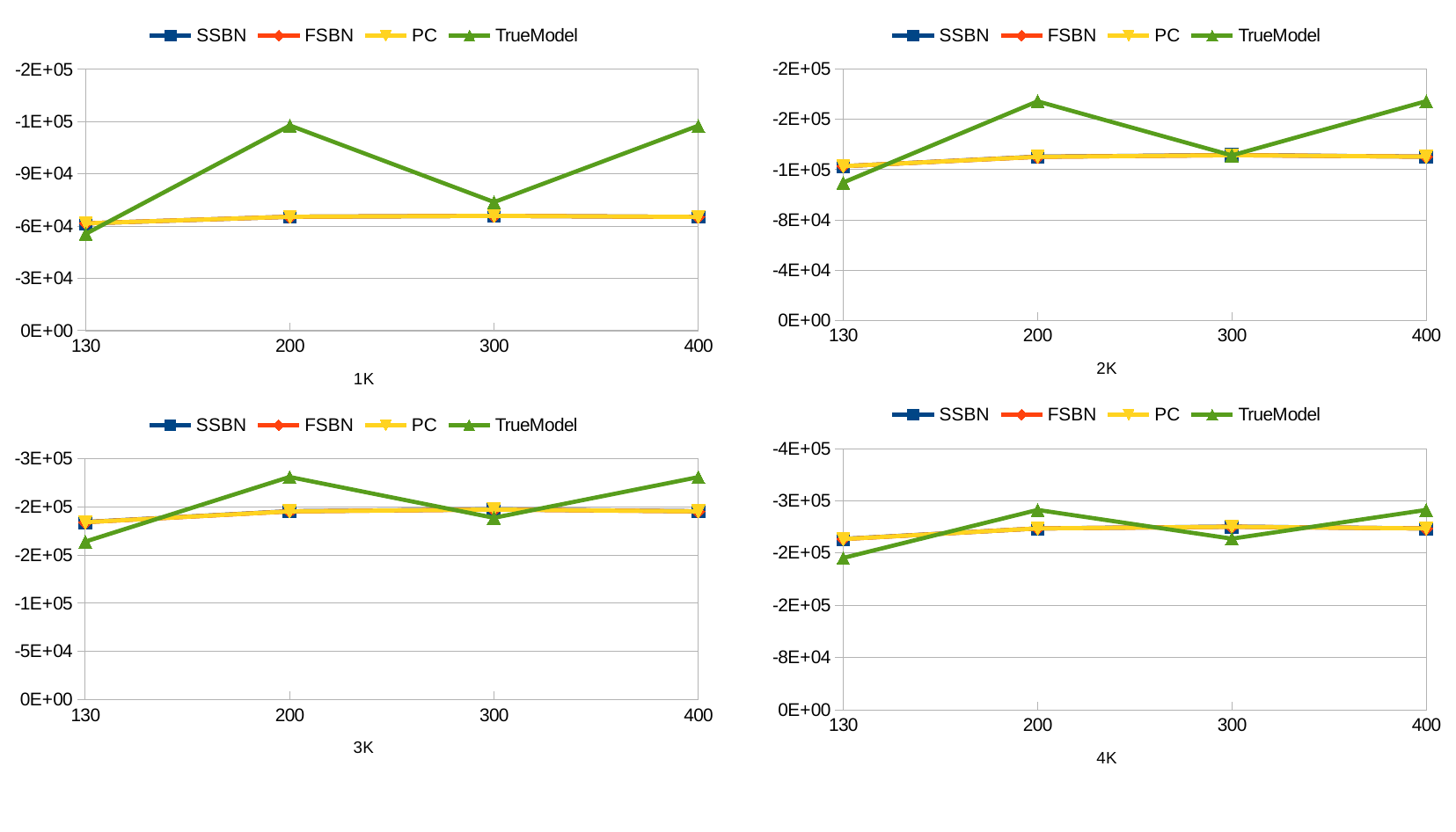}\\
  \caption{Quality of density estimation (\emph{BDeu score}) about PC, FSBN, and SSBN and true networks with various densities (Note that the values on the Y-axis vary for each graph in this figure).}\label{figd_den_bdeu}
\end{figure}

\begin{figure}[htbp]
  \centering
  \includegraphics[scale=0.28]{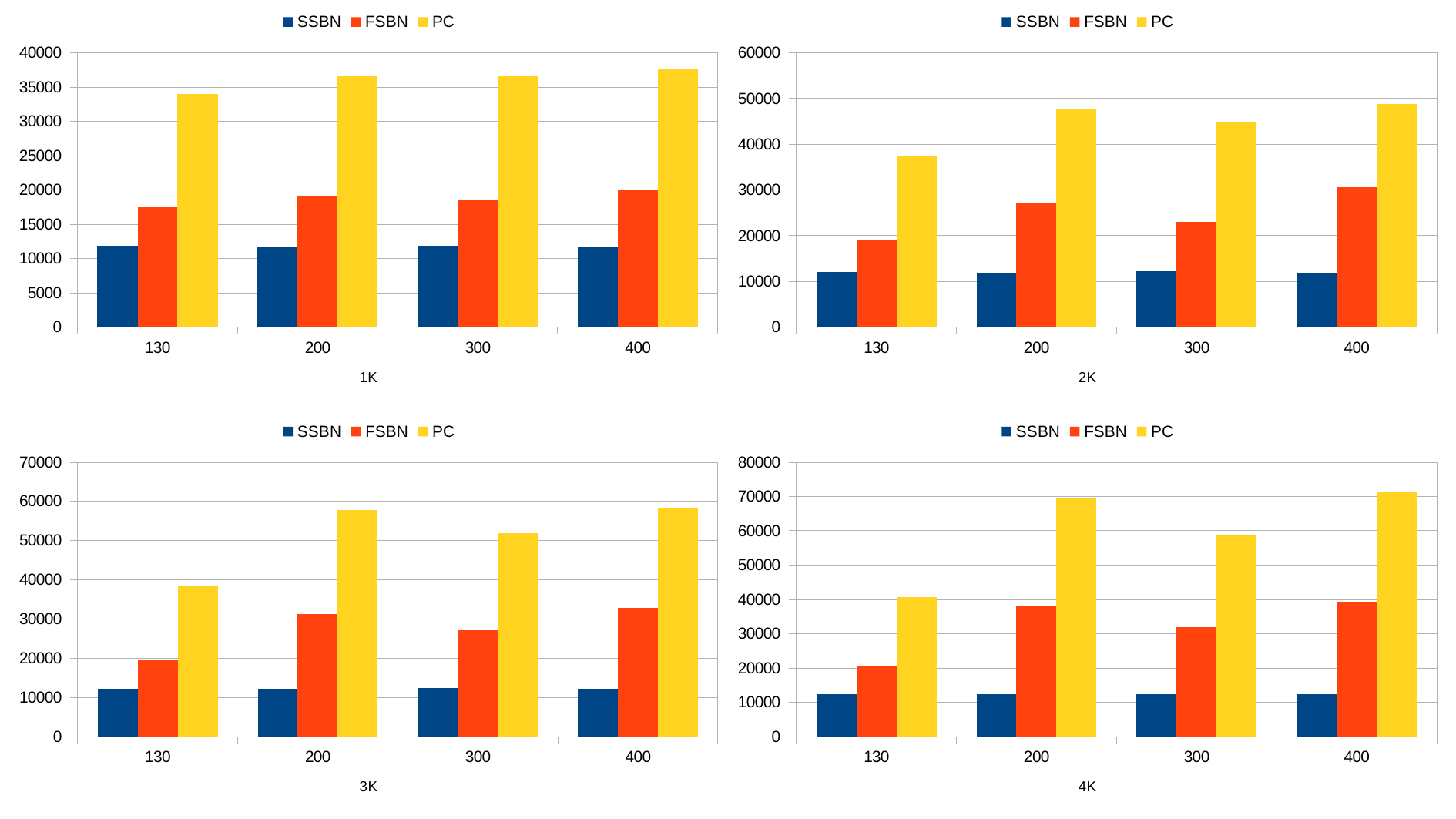}\\
  \caption{Computational complexity of PC, FSBN and SSBN to induce given networks with various densities (Note that the values on the Y-axis vary for each graph in this figure).}\label{figd_den_wighted}
\end{figure}

In the experiment, a synthetically produced 100 nodes network was used with a fixed number of nodes but varying numbers of edges to simulate different network densities (as shown in Table \ref{tb_data_density}). The results are presented in Figures~\ref{figd_den_hamming},~\ref{figd_den_bdeu}, and \ref{figd_den_wighted} for the comparison of knowledge discovery, density estimation, and time efficiency.

\begin{table}[h]
\caption{Description of data sets}\label{tb_data_density}
\centering
\begin{tabular}{lccc}
  \hline
  Data set(s)  & \# Instances & \# Nodes & \# Arcs\\
  \hline
  100-I & 1K to 4K & 100 & 130 \\
  100-II & 1K to 4K  & 100 & 200 \\
  100-III & 1K to 4K  & 100 & 300 \\
  100-IV & 1K to 4K  & 100 & 400\\

  \hline\hline
\end{tabular}
\end{table}

The following observations can be made:
\begin{enumerate}
    \item All three algorithms (PC, FSBN, and SSBN) maintain a consistent level of learning quality even as the network density increases. This indicates that the proposed heuristics in FSBN and SSBN are effective in discovering the network structure, and they perform well in capturing the dependencies between variables.
    \item Similarly, all three algorithms demonstrate comparable performance in density estimation as the network density increases. This suggests that the learning quality of the algorithms is not significantly affected by the density of the network.
    \item As expected, all three algorithms require more CI tests as the number of edges increases in denser networks. However, SSBN exhibits a much slower growth rate in the number of CI tests compared to PC and FSBN. This indicates that SSBN is more efficient in terms of time consumption, particularly in denser networks.
\end{enumerate}

Overall, the results indicate that all three algorithms (PC, FSBN, and SSBN) maintain consistent learning quality as the network density increases. However, SSBN demonstrates superior time efficiency compared to PC and FSBN, making it a promising approach for learning the structure of denser networks.

\section{Conclusion}\label{sec:con}
The novel search heuristics proposed in FSBN and SSBN utilize the topology information obtained from previously executed CI tests to guide future searches. These algorithms have been proven to be sound and effective. Experimental studies demonstrate that they can achieve comparable results in terms of learning quality as the classical PC algorithm but with significantly reduced computational resources. SSBN, in particular, shows even greater efficiency than FSBN, especially in denser networks, indicating its potential for solving larger-scale problems.

FSBN and SSBN highlight the untapped potential for enhancing traditional constraint learning algorithms by leveraging the underlying topology information. Future research in this area could focus on better utilization of such information. For instance, the RDSA of a node $X$ showing a relative degree of separation of ancestors, determined during the search for its neighbors, is currently not utilized in the study of other nodes' neighbors. If methods can be developed to transfer and incorporate this type of information, further improvements in time efficiency can be achieved.

Thus, the FSBN and SSBN algorithms have great potential for building causal local surrogate models for black-box models and detecting concept/data drift in the era of big data. Future research can investigate techniques that target the reduction of computational burden in diverse application domains. This can be achieved by focusing on the following directions: Causal Local Surrogate Models: Black-box models, such as deep neural networks, often lack interpretability, making it challenging to understand the underlying causal relationships and provide explanations with the help of Markov blanket for their predictions~\cite{ribeiro2016should,vu2020pgm,minn2023laplace}. FSBN and SSBN can assist in building causal local surrogate models that approximate the behavior of the black-box model in a local region of the input space. These surrogate models, represented as BNs, capture the causal relationships among variables and provide interpretable explanations for the predictions made by the black-box model. They allow us to understand how changes in the input variables affect the output and provide insights into the decision-making process of the black-box model. Concept/Data Drift Detection: In the era of big data, where the underlying data distribution may change over time, it is crucial to detect concepts or data drift quickly. Concept drift refers to the change in the underlying data-generating process, while data drift refers to changes in the data distribution itself. FSBN and SSBN can be applied to monitor the drift in the causal relationships among variables. By periodically updating the learned structure of the BNs and comparing it with the previous models, deviations or changes in the causal relationships can be detected.

 Overall, both algorithms of FSBN and SSBN have a high potential to assist in building causal local surrogate models that can enhance our understanding of complex black-box models, provide interpretability, and enable the detection of changes in causal relationships over time. Those methods can also help to identify when the black-box model's assumptions are no longer valid or when the data distribution has significantly changed, allowing for timely adaptation and retraining of the models. This is particularly valuable in domains where interpretability, trustworthiness, and adaptability to evolving data are critical, such as healthcare, finance, and autonomous systems.

\bibliographystyle{IEEEtran}
\bibliography{IEEEabrv,biblio}
\end{document}